# On the Detection of Conflicts in Diagnostic Bayesian Networks Using Abstraction

Young-Gyun Kim
Department of Computer Science
University of South Carolina
Columbia, SC 29208
< ykim@usceast.cs.sc.edu >

Marco Valtorta
Department of Computer Science
University of South Carolina
Columbia, SC 29208
< mgv@usceast.cs.sc.edu >

## Abstract

An important issue in the use of expert systems is the so-called brittleness problem. Expert systems model only a limited part of the world. While the explicit management of uncertainty in expert systems mitigates the brittleness problem, it is still possible for a system to be used, unwittingly, in ways that the system is not prepared to address. Such a situation may be detected by the method of straw models, first presented by Jensen et al. [1990] and later generalized and justified by Laskey [1991]. We describe an algorithm, which we have implemented, that takes as input an annotated diagnostic Bayesian network (the base model) and constructs, without assistance, a bipartite network to be used as a straw model. We show that in some cases this straw model is better that the independent straw model of Jensen et al., the only other straw model for which a construction algorithm has been designed and implemented.

## 1 INTRODUCTION

Developing a diagnostic expert system that uses a Bayesian network as a model of the domain of interest requires several activities, such as knowledge acquisition, learning, sensitivity analysis, and conflict checking [Andreassen et al., 1987; Andersen et al., 1989; Spiegelhalter et al., 1993]. By *conflict* we mean that the model is no longer valid for the given data [Jensen et. al., 1990; Laskey, 1991][1]. In other words, for a given piece of data, we cannot have any confidence in the results coming from the system. Note that this is possible even though Bayesian networks cannot be inherently inconsistent, in the sense that, if their (local) conditional probability tables are consistent, they always encode a probability distribution [Pearl, 1988, Chapter 3; Neapolitan, 1990, Theorem 5.2].

---

[1]The definition of conflict in this paper is not that used in the model-based diagnosis literature, where a conflict is defined as a set of components at least one of which must be faulty.

In practice, what can happen is that, while using a Bayesian network in a diagnostic application, we observe that the evidence on which the diagnosis should be based is very unlikely given the network. It may be that the evidence at hand describes a rare case that the network is fully qualified to handle, but it may also be that the network is not equipped to handle the case, and that the low probability of the case is an indication of that. We maintain that it is important for Bayesian network to distinguish between the two situations just described.

Consider, as a simple example, the use of a system designed to classify geological resources. This system, when presented with data from, say, the moon, would show the data to be very unlikely. A (naive) use of the system would wonder whether this low probability derives from the presence of an unusual sample that the system is built to handle, or whether the system is simply not built to handle the sample. In the example, the latter situation holds. By definition, a geological resource is a resource of the earth, but our hypothetical user's command of the English (or Greek) language is weak enough. It may be disastrous or very expensive to treat the sample as correctly classified. It would be very desirable for the system to indicate to the user the existence of a conflict.

The world we live in is filled with countless factors. If we restrict our interest to the medical world, some factors might be age, gender, stress, and so on. If we have a disease, we try to find out why it happened and how to cure it. In other words, the main concern in this world is to find out relationships among factors. Hence, we need a world that has a probability distribution that represents the relationships among factors. We call this world a model.

Because the world is big and complex, models of the world are difficult to design and represent. We use instead a small world in which we ignore some factors and some relationship, assumed to be small and unimportant. Every world we can express and use is to be considered a small world. Moreover, the models we can manage are always approximated models for a small world, obtained by making some background assumptions. In certain conditions, called *conflict data*, that violate the background assumptions, the model may be weak. If the model is presented with conflict data, the results obtained from it are unreliable.

When we observe that the probability of some data is very low in a model, it is difficult to decide whether the low probability comes from a rare situation covered in the model or from data conflict. The best way to solve this problem is to compare the result with the full world model [Laskey and Lehner, 1992; 1994]. Since it is impossible to construct this kind of model, we fall back on the next best option, which is to construct a better model and compare results. But the reason for checking conflict is not only to diagnose the model itself, but also to trigger revision for a better model. In other words, we cannot assume the availability of a better model; if we had it, we would use it instead of the model for which a conflict is suspected. To overcome this quandary, Laskey [1991] suggests *straw models* that "capture some of the expert's intuitions about how the model could go wrong, but are computationally much simpler than the fully specified alternate model." In particular, a straw model may be an approximated model of a given model. The straw model is structurally poor and less probable than the given model because of the absence of *details* (factors and relationships) that may be very relevant. However, the absence of these factors and relationships can make the straw model more probable when some conflict is caused by them.

By comparing a given model with a straw model, a conflict can be detected easily and automatically without construction of a better model that requires more effort. More concretely, suppose that we obtain a low probability of some findings (evidence) that we suspect to be too low to be happening in a real-life situation. Further suppose that the probability of the findings in the straw model is higher than the probability of the findings in the given model. Then, we can conclude very reliably that the evidence is a conflict for the given model, and we should alert the user of this.

Laskey [1991] quantifies these issues as follows. (Also see Laskey [1994] and Laskey and Lehner [1992; 1994] for related material.) A *world* is a vector of propositional variables and a *model* can be represented by assigning a probability distribution over the variables in the world. Assume that a small world $\underline{X}$ is embedded within a larger world $\underline{W} = (\underline{X'}, \underline{Y})$. We use the following notation.

$\underline{X}$: A vector of propositional variables that represents a small world. Each variable $X_i$ can take on values in a set $x_{i1}, \ldots, x_{ik}$.

$\underline{X'}$: A vector with the same variables as $\underline{X}$, but which has additional outcomes in the larger world that are not represented in the small world.

$\underline{Y}$: A vector of variables that are not explicitly represented in the small world.

$\underline{x}$: Variables to which a value is assigned (*data* or *evidence* variables).

$P^a(\cdot)$: Probability distribution encoded in the model (usually, assessed by the expert).

$P^o(\cdot)$: Probability distribution of alternate (straw) model.

$P(\cdot)$: Probability distribution on the larger world that $P^a$ intends to approximate.

Using the notation defined above, the global probability distribution $P(\cdot)$ is what we want to approximate with the distribution $P^a(\cdot)$. The following statements and theorem explain the relation between these two probability distributions and how a straw model (with its distribution $P^o(\cdot)$) can be used to detect conflict.

Let the proposition $q$ represent the background assumptions for $P^a(\cdot)$:

$$q = (\bigwedge_i X_i' \in \{x_{i1}, \ldots, x_{ik}\}) \wedge (\bigvee_{\underline{y}_j} Y = \underline{y}_j)$$

For any $\underline{x}$ in the small world $\underline{X}$, we have:

$$P^a(\cdot) = P(\underline{x} \mid q) = \sum_{\underline{y}} P(\underline{x}, \underline{y} \mid q)$$

When the model $P^a(\cdot)$ is not appropriate, we have:

$$P^o(\cdot) = P(\underline{x} \mid \neg q) = \sum_{\underline{y}} P(\underline{x}, \underline{y} \mid \neg q)$$

Let $P(q) = 1 - \epsilon$. The model $P(\cdot)$, restricted to the variables $\underline{X'}$, can be written:

$$P(\underline{x}) = (1 - \epsilon)P^a(\underline{x}) + \epsilon P^o(\underline{x})$$

**Theorem 1 (Surprise Index) [Laskey, 1991].** *Let $P^a(\cdot)$ and $P^s(\cdot)$ be probability distributions over $\underline{X}$. Define the index of surprise at evidence $\underline{x}_e$ under $P^a(\cdot)$ relative to $P^s(\cdot)$ as:*

$$c_s = \log[P^s(\underline{x}_e)/P^a(\underline{x}_e)]$$

*Let $\pi_k$ be the probability under $P^a$ that $c_s$ is greater than $K$. Then $\pi_k = 2^{-K}$.*

When $P^a(\cdot)$ applies, $P^s$ should be much more probable than $P^a$ to produce high values of the surprise index. Equivalently, "high values of conflict are a priori unlikely when the assessed model is considered probable. In other words, any alternate model specified a priori is unlikely to fit the data much better than the assessed model" [Laskey, 1991, p. 201]. It is therefore reasonable to inform the user that a conflict is possible when the surprise index has a high value.

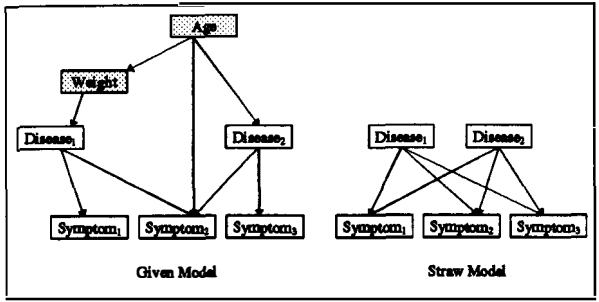

Figure 1: Given model and bipartite straw model.

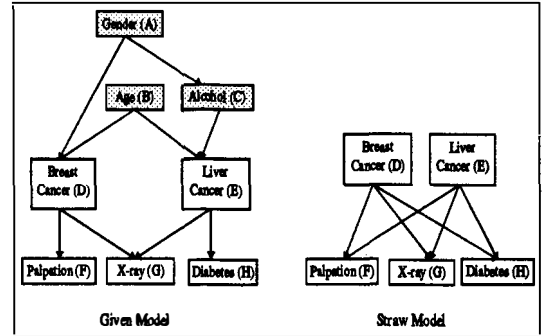

Figure 2: The diagnosis system for liver and breast cancer.

## 2   THE BIPARTITE STRAW MODEL

A bipartite straw model can be obtained by the elimination of some factors (variables) from the given model[2]. Elimination of variables is a classic kind of abstraction. (See, e.g., [Mozetic, 1991].) Clearly, the quality of the straw model depends critically on which variables are eliminated, a choice that depends on the domain and on the task addressed by the expert system. Since we are interested in diagnosis, we consider the task of heuristic classification and divide variables into three groups. One is called *Target* and consists of the variables for which we want to know the probabilities. The next group is *Evidence*, and the findings belong to this group. The last group is called *Other*, and it includes variables that do not belong to the above two groups. In medical diagnosis the target variables might be diseases, the evidence variables may by symptoms, and the variables in *Other* may be factors such as age or weight[3]. (See Figure 1 for an illustration.)

Since we construct the bipartite straw model by eliminating the *Other* variables, the straw model consists of the target and evidence variables only. More precisely and formally, letting $pa(A)$ be the parents of node $A$, we have:

$$U = Target \cup Evidence \cup Other$$

---

[2]We could have called this straw model *naive Bayes* or *idiot Bayes* model, but we chose a more neutral name for it.

[3]While our implementation of the algorithm for construction of the bipartite straw model does not assume any relative ordering of the three groups of variables, in the diagnostic applications we have in mind the target variables precede the evidence variables, and the *Other* variables either are setting factors that precede the target variables, or are physiological states that are placed between the target variables and the evidence variables.

$$U^{straw} = Target \cup Evidence$$
$$pa(A) = Target \quad \forall A \in Evidence$$

Then, for each $A \in Evidence$ and for each $\underline{Target}$, the vector of $Target$ variables, we have:

$$P^{straw}(A \mid \underline{Target}) = \sum_{\backslash A} P^{given}(A \mid finding = \underline{Target})$$

We can therefore obtain the entries in the table of conditional probabilities for each evidence variable by belief propagation. Similarly, for all $A \in Target$, the prior probabilities for the straw model are related to probabilities in the given model as follows:

$$P^{straw}(A) = \sum_{\backslash A} P^{given}(A \mid finding = \{\})$$

The details of the algorithm used to compute $P^{straw}$ are beyond the scope of this paper and can be found in [Kim, 1994].

In our bipartite straw model, all target variables are parents of each evidence variable, and the target variables are conditionally independent of each other (cf. Figure 1). The variable elimination method makes the straw model weak and structurally poor when the *Other* variables give a high value of probabilistic causation, such as, for example, in the case of old age and neuralgia. But when conflict data is entered, and the conflict is caused by variables in *Other*, the bipartite straw model is more probable than the given model.

## 3   EXAMPLE OF CONSTRUCTION AND USE OF THE BIPARTITE STRAW MODEL

Assume a (fictitious) medical diagnosis system for liver and breast cancer represented by the Bayesian network

|  | Age (B) | |
|---|---|---|
|  | below 30 | above 30 |
| Gender (A) = male | (0, 100) | (1, 99) |
| Gender (A) = female | (20, 80) | (50, 50) |

Joint probability table for
Breast Cancer given Gender and Age

|  | Breast Cancer (D) | |
|---|---|---|
|  | yes | no |
| Palpation (F) | (95, 5) | (5, 95) |

Conditional probability table for
Palpation given Breast Cancer

| Gender (A) = (male, female) | (50,50) |
|---|---|
| Age (B) = (below 30, above 30) | (20,80) |

Prior probability table for
Gender and Age

|  | Liver Cancer (E) | |
|---|---|---|
|  | yes | no |
| Diabetes (H) | (95, 5) | (5, 95) |

Conditional probability table for
Diabetes given Liver Cancer

|  | Breast Cancer (D) | |
|---|---|---|
|  | yes | no |
| Liver Cancer (E) = yes | (95, 5) | (80, 20) |
| Liver Cancer (E) = no | (80, 20) | (5, 95) |

Joint probability table for
X-ray given Breast and Liver Cancer

|  | Gender (A) | |
|---|---|---|
|  | male | female |
| Alcohol (B) | (95, 5) | (10, 90) |

Conditional probability table for
Alcohol given Gender

|  | Age (B) | |
|---|---|---|
|  | below 30 | above 30 |
| Alcohol (B) = yes | (10, 90) | (40, 60) |
| Alcohol (B) = no | (2, 98) | (10, 90) |

Joint probability table for
Liver Cancer given Alcohol and Age

Figure 3: Probability tables for a fictitious diagnostic system.

of Figure 2 with conditional probability tables given in Figure 3.

The given model is such that palpation is associated with females and diabetes is associated with males. The findings (Palpation = yes, Diabetes= no) conflict with the assumptions used when building he model. This is well detected by a positive conflict index. Our program gives the following results:

$$P^{given}(Palpation = yes, Diabetes = yes) = 0.0452$$

$$P^{straw}(Palpation = yes, Diabetes = yes) = 0.0619$$

Therefore the conflict index $c_s = \log(P^{straw}/P^{given})$ is positive and the user is alerted of a conflict.

## 4   COMPARISON OF THE INDEPENDENT AND BIPARTITE STRAW MODELS

Finn Jensen et al. [1990] propose a straw model in which all variables are independent of each other and have implemented this straw model to compute the conflict index in HUGIN. Figure 4 illustrates the independent straw model. Jensen and his collaborators assume that, in the absence of conflict, the joint probability of all evidence variables is greater that the product of probabilities of each evidence variable when the given model is applied, i.e., $P(x, y) > P(x)P(y)$. They argue that this is normally the case, because $P(x|y) > P(x)$ and $P(x, y) = P(x|y)P(y)$. For the independent straw model, the conflict measure is defined as:

$$conf(x, \ldots, y) = \log[(P(x) \times \ldots \times P(y))/P(x, \ldots, y)],$$

where $x, \ldots, y$ are the findings.

The independent straw model is computationally simpler than the bipartite straw model, but it may fail to detect conflict when the above assumption is violated, as shown in the following example, which continues the example of the previous section.

If we obtain findings (Palpation = yes, X-ray = yes, Diabetes = yes) in the model of Figure 2, the relevant probabilities are:

$$P^{given}(Palpation = yes) = 0.252$$

$$P^{given}(X - ray = yes) = 0.365$$

$$P^{given}(Diabetes = yes) = 0.247$$

$$P^{straw}(Palpation = yes, X - ray = yes,$$
$$Diabetes = yes) = 0.252 \times 0.365 \times 0.247 = 0.0227$$

$$P^{given}(Palpation = yes, X - ray = yes,$$
$$Diabetes = yes) = 0.0388$$

Hence, the conflict index $c_s$ is

$$c_s = \log(0.0227/0.0388)$$

which is negative and indicates absence of conflict.

On the other hand, since using the bipartite straw model $P^{straw}(Palpation = yes, X - ray = yes, Diabetes = yes) = 0.0551$, the conflict index using the bipartite straw model is

$$c_s = \log(0.0551/0.0388)$$

which is positive and indicates presence of conflict.

We emphasize that a positive conflict index does not necessarily indicate that the given model is bad; it only alerts the user that assumptions taken in constructing the given model may not be appropriate for the data at hand. In the example, the given model was built with the background assumption that diabetes is associated with males and palpation is associated with females. This assumption is not satisfied in the case at hand.

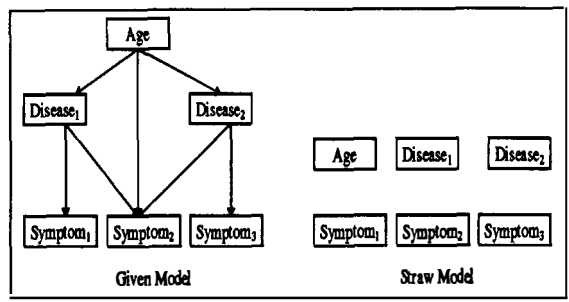

Figure 4: Independent straw model.

# 5 CONCLUSIONS

We have shown that the proposed bipartite straw model can be constructed automatically by eliminating the variables in *Other*, as described in the Section entitled "The Bipartite Straw Model." This construction requires several procedures, which we implemented in the C language under the ULTRIX operating system. The two main algorithms used for the computation of the probabilities needed to obtain the conflict index are Lauritzen and Spiegelhalter's algorithm [1988], as described in [Neapolitan, 1990] and the algorithm by F.V. Jensen, described in [Jensen, 1990] The details of the system we have implemented are given in [Kim, 1994].

The main contributions of our research may be summarized as follows:

1. We have demonstrated that it is possible to construct straw models different and sometimes better than the ones that have been presented so far in the literature (independent-type straw models)

2. These new models (which we call bipartite straw models) can be constructed by abstraction from diagnostic Bayesian networks (heuristic classification models) with minimal intervention from knowledge engineers or experts, who only need to identify the target and evidence nodes in their models.

3. To substantiate the claims, we wrote a program and tested it over several examples.

Here is a list of suggestions for further research.

1. The bipartite straw model is only one of various possible ones. There may be other methods for designing straw models through various other elimination methods or the use of other (automatic or semi-automatic) approximations and abstractions, such as, e.g.:

- Eliminate variables for which one value is much more probable than the others.
- Eliminate certain states in variables [Wellman, 1994].
- Combine certain variables into one.

2. This approach should be tested on more realistic models.

3. Conditions under which bipartite straw models are better than independent straw models should be investigated. We expect he bipartite straw model to be more accurate than the independent straw model, since it is structurally more complex. This, however, is not necessarily true in all situations, because the bipartite straw model drops the *Other* variables, such as Age in the example of Figure 2.

4. The conflict (surprise) index should be characterized as a measure, if possible.

5. The complexity of computing the conditional probability tables for the straw models needs to be analyzed. In particular, conditions under which the computation of the new tables can be done in polynomial time need to be investigated.

## Acknowledgments

We are thankful to Professors Finn V. Jensen, John Rose, Juan E. Vargas, and the anonymous referees for their helpful comments. After the paper was reviewed, the authors became aware of Michael Wellman's work on abstraction in Bayesian networks; we thank Professor Wellman for general discussions on abstraction. M.V. gratefully acknowledges support from the U.S. Department of Agriculture for the project "Expert Systems for Agricultural Loans: Collaboration with S.C. State University."